\documentclass{article}

\usepackage{PRIMEarxiv}

\usepackage[utf8]{inputenc} 
\usepackage[T1]{fontenc}    
\usepackage{hyperref}       
\usepackage{url}            
\usepackage{booktabs}       
\usepackage{amsfonts}       
\usepackage{nicefrac}       
\usepackage{microtype}      
\usepackage{lipsum}
\usepackage{fancyhdr}       
\usepackage{graphicx}       
\graphicspath{{media/}}     
\usepackage{multirow}

\title{Img2Loc: Revisiting Image Geolocalization using Multi-modality Foundation Models and Image-based Retrieval-Augmented Generation
}

\author{
  Zhongliang Zhou$^*$ \\
  University of Georgia \\
  \texttt{zz42551@uga.edu} \\
   \And
  Jielu Zhang$^*$ \\
  University of Georgia \\
  \texttt{jz20582@uga.edu} \\
   \And
  Zihan Guan \\
  University of Virginia \\
  \texttt{bxv6gs@virginia.edu} \\
   \And
  Mengxuan Hu \\
  University of Virginia \\
  \texttt{qtq7su@virginia.edu} \\
   \AND
  Ni Lao \\
  University of Georgia \\
  \texttt{noon99@gmail.com} \\
   \And
  Lan Mu \\
  University of Georgia \\
  \texttt{mulan@uga.edu} \\
   \And
  Sheng Li$^{\dagger}$ \\
  University of Virginia \\
  \texttt{shengli@virginia.edu} \\
   \And
  Gengchen Mai$^{\dagger}$ \\
  University of Georgia \\
  \texttt{gengchen.mai25@uga.edu} \\
}

\begin{document}
\maketitle
\def\thefootnote{*}\footnotetext{These authors contributed equally to this work}
\def\thefootnote{$\dagger$}\footnotetext{Corresponding authors}

\begin{abstract}
Geolocating precise locations from images presents a challenging problem in computer vision and information retrieval. Traditional methods typically employ either classification—dividing the Earth's surface into grid cells and classifying images accordingly, or retrieval—identifying locations by matching images with a database of image-location pairs. However, classification-based approaches are limited by the cell size and cannot yield precise predictions, while retrieval-based systems usually suffer from poor search quality and inadequate coverage of the global landscape at varied scale and aggregation levels. To overcome these drawbacks, we present \textbf{Img2Loc}, a novel system that redefines image geolocalization as a text generation task. This is achieved using cutting-edge large multi-modality models (LMMs) like GPT-4V or LLaVA with retrieval augmented generation. Img2Loc first employs CLIP-based representations to generate an image-based coordinate query database. It then uniquely combines query results with images itself, forming elaborate prompts customized for LMMs. When tested on benchmark datasets such as Im2GPS3k and YFCC4k, Img2Loc not only surpasses the performance of previous state-of-the-art models but does so without any model training.
\end{abstract}


\section{Introduction}

The field of visual recognition has witnessed a marked improvement, with state-of-the-art models significantly advancing in areas such as object classification \cite{sumbul2019bigearthnet,cong2022satmae,mai2023sphere2vec,mai2023csp}, object detection \cite{zhang2022dino, carion2020end,li2023rethink}, semantic segmentation \cite{wang2024mixsegnet, wang2023densely, wang2023dual, zhang2023text2seg}, scene parsing \cite{hou2020strip, li2020semantic}, disaster response \cite{he2022earth, he2021deep}, environmental monitoring\cite{he2022quantifying} among others \cite{chen2023ride, chen2023distributional, chen2023real}. As progress moves forward, the information retrieval community is widening its focus to include the prediction of more detailed and intricate attributes of information. A key attribute in this expanded scope is image geolocalization \cite{hays2008im2gps, chen2020survey,cepeda2023geoclip}, which aims to determine the exact geographic coordinates given an image. The ability to accurately geolocalize images is crucial, as it provides possibilities for deducing a wide array of related attributes, such as temperature, elevation, crime rate, population density, and income level, providing a comprehensive insight into the context surrounding the image.

\begin{figure}[h]
  \centering
  \includegraphics[width=0.5\columnwidth]{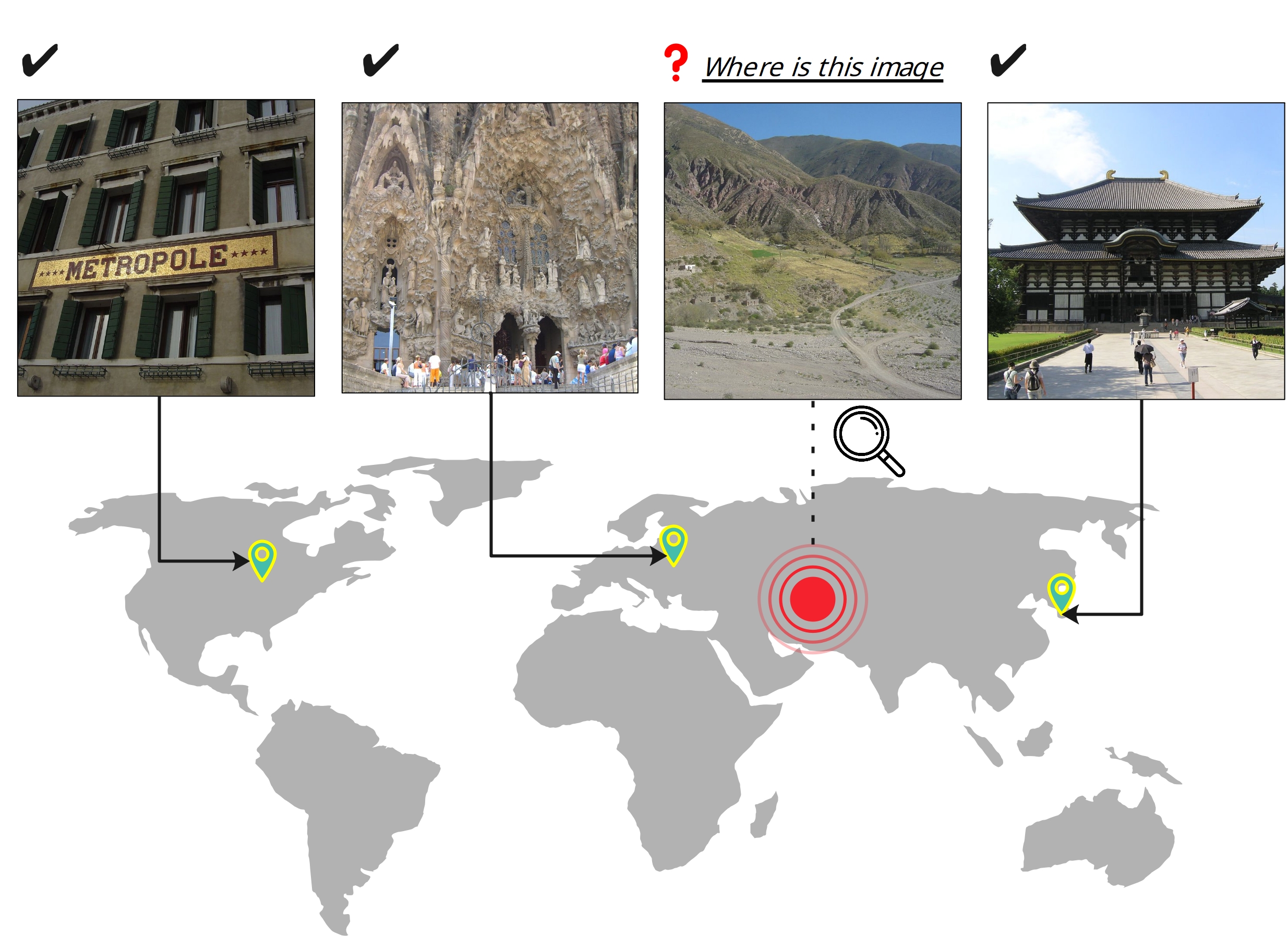}
  \caption{The image geolocalization problem refers to predicting the coordinates of any given image.}
  \label{figure1}
\end{figure}

In our study, we delve into predicting the geographic coordinates of a photograph solely from the ground-view image. Predictions are considered accurate if they closely match the actual location (Figure \ref{figure1}). Prevailing research approaches fall under the categories of either retrieval-based or classification-based methods. Retrieval-based techniques compare query images against a geo-tagged image database \cite{workman2015wide, liu2019lending, zhu2021vigor, yang2021cross, zhu2022transgeo, tian2017cross, shi2020looking, zhu2023r2former}, using the location of the image that closest matches the query image to infer its location. Although straightforward, this method faces hurdles such as the complexity of feature extraction, the computational intensity of nearest neighbor search, and potential inaccuracies from over-reliance on database image locations. Alternatively, classification-based methods treat geolocalization as a classification task \cite{Pramanick22, vo2017revisiting, muller2018geolocation, clark2023we, weyand2016planet, seo2018cplanet, izbicki2020exploiting}, segmenting the Earth into discrete cells and training neural networks to classify images into these cells. However, this approach can yield significant errors, especially when the actual location of an image differs significantly from the center of its assigned cell. Moreover, the predefined cell structure introduces inherent limitations and biases, decreasing generalizability and accuracy across various global locations.

Given the inherent limitations of both retrieval and classification approaches, we are transitioning to the more recent and increasingly dominant approach of foundation models. In this vein, we propose a generative approach that predicts the geographic coordinates of new query images using a reference gallery and multimodality language models, named \textbf{Img2Loc}. Initially, we transform all geo-tagged images into embeddings using the CLIP model \cite{radford2021learning}, creating a vast embedding space. To navigate this space efficiently, we use a vector database with GPU-accelerated search algorithms, quickly pinpointing and retrieving gallery images similar to the query images. Next, we formulate elaborate prompts integrating the image and the geographical coordinates of these reference points and feed them into state-of-the-art multi-modality models like GPT-4V~\cite{yang2023dawn} or LLaVA ~\cite{liu2023visual}, known for their adeptness in generating accurate outputs from combined image and text inputs. To further improve the accuracy, we introduce negative sampling by identifying and using the most dissimilar points in the database as a counterreference. This sharpens the model's ability to distinguish between relevant and irrelevant data points. 
Our model, when evaluated on well-established datasets such as Im2GPS3k \cite{vo2017revisiting} and YFCC4k \cite{vo2017revisiting}, demonstrates notable advances, significantly outperforming the prior state-of-the-art methods without any model fine-tuning. This highlights the efficacy of our generative approach, which synergizes the retrieval method's strengths with the advanced understanding and generative prowess of contemporary language models.

In summary, our study makes significant strides for the task of image geolocalization, marked by the following  contributions:
\begin{itemize}
    \item To the best of our knowledge, this study is the first successful demonstration of multi-modality foundation models in addressing the challenges of geolocalization tasks.
    \item Our approach is training-free, avoiding the need for specialized model architectures and training paradigms and significantly reducing the computational overhead.
    \item Using a refined sampling process, our method not only identifies reference points closely associated with the query image but also effectively minimizes the likelihood of generating coordinates that are significantly inaccurate.
    \item We achieve outstanding performance on challenging benchmark datasets including Im2GPS3k and YFCC4k compared with other state-of-the-art approaches.
\end{itemize}

\section{Related Work}

\paragraph{\textbf{Image Geolocalization as a classification task.}} The predominant approach for the image geolocalization problem involves first segmenting the planet's surface into discrete grids, such as the Google S2 grid, and assigning a geographic coordinate to each grid \cite{Pramanick22, vo2017revisiting, muller2018geolocation, clark2023we, weyand2016planet, izbicki2020exploiting, seo2018cplanet}. This methodology permits a model to directly predict a class, thereby simplifying the complex task of geolocalization into a more manageable form of classification. To refine this approach and introduce granularity into the prediction, recent advances have involved partitioning the Earth's surface into multiple levels, offering a hierarchical, multi-scale perspective of localization \cite{muller2018geolocation}. However, while this cell-based classification system simplifies the prediction process, it inherently introduces localization errors, particularly if the actual location of interest lies far from the center of the predicted cell. This discrepancy stems from the coarse nature of cell-based classification, where the precision of localization is inherently limited by the size and scale of the cells defined in the model. 

\paragraph{\textbf{Image Geolocalization as a retrieval task.} }
In another direction, image geolocalization has significantly evolved from rudimentary methods to sophisticated retrieval-based systems over the years \cite{workman2015wide, liu2019lending, zhu2021vigor, yang2021cross, zhu2022transgeo, tian2017cross, shi2020looking, zhu2023r2former}. Retrieval-based systems, recognized for their intuitiveness, leverage these extensive databases to find matches for query images based on feature similarity in a multi-dimensional space. However, creating planet-level reference datasets for these systems presents formidable challenges, not limited to scale but also encompassing data diversity, temporal changes, and the need for precise annotations. To address the intrinsic differences in ground and aerial perspectives, separate models are often adopted, with the integration of these models aiming to provide a more comprehensive understanding and robust localization system \cite{regmi2019bridging}. Nonetheless, this integration introduces the significant hurdle of misalignment between perspectives, potentially undermining accuracy. An innovative solution to this challenge is the implementation of non-uniform cropping \cite{zhu2022transgeo}, which selectively focuses on the most informative patches of aerial images. This method prioritizes features that offer distinctive geographical cues, enhancing precision by addressing the issue of non-uniform feature distribution across different views.

\begin{figure*}[!ht]
  \centering
  \includegraphics[width=\textwidth]{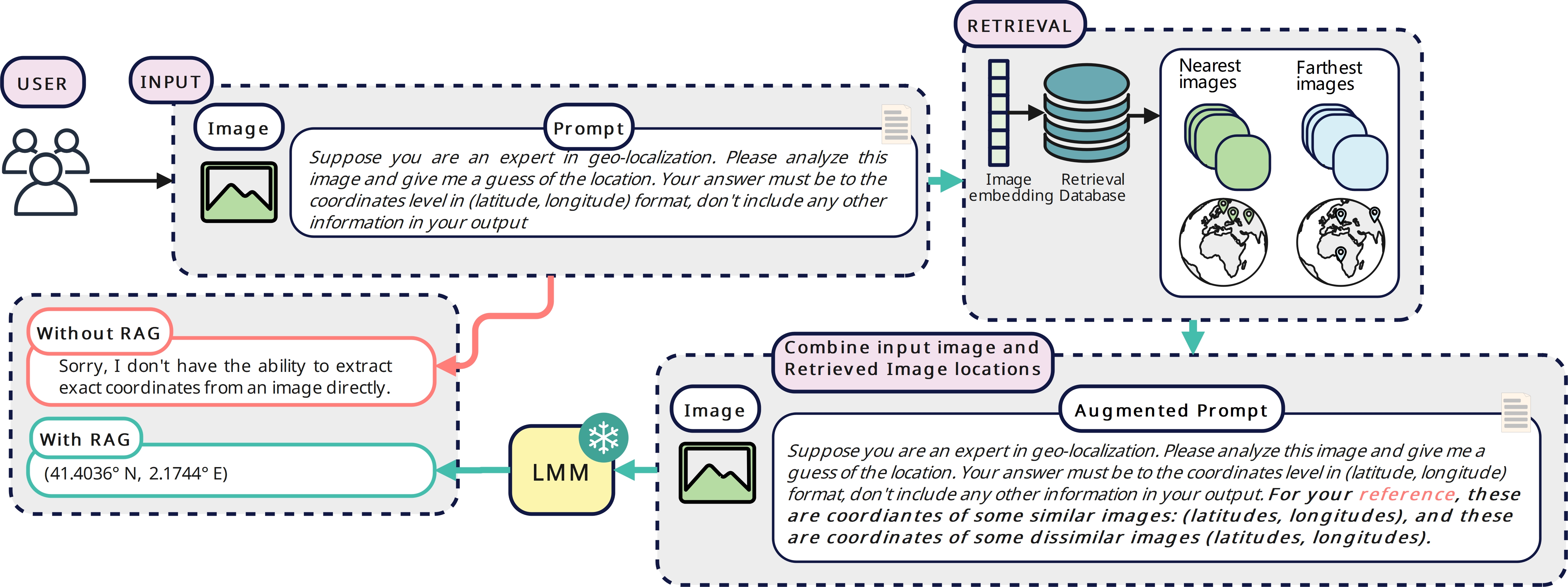}
  \caption{The architecture of the proposed framework.}
  \label{figure3}
\end{figure*}

\paragraph{\textbf{Multi-modality foundation models and Retrieval-Augmented Generation.}} Large language models like GPT-4 \cite{achiam2023gpt} and LLaMA \cite{touvron2023llama} have set new benchmarks in natural language processing, surpassing human performance in a variety of tasks as evidenced by their results on SuperGLUE \cite{wang2019superglue} or BIG-bench \cite{srivastava2022beyond}. Their exceptional zero-shot capabilities enable these generative models to be applicable across a wide range of research domains. Building on this success, multi-modality models such as GPT-4V \cite{yang2023dawn} and LLaVA \cite{liu2023visual} have extended the prowess of large language models (LLMs) into the visual domain. However, alongside these remarkable achievements, certain challenges have become evident, most notably issues related to hallucination and reliance on outdated knowledge databases. These issues can compromise the reliability and trustworthiness of models' outputs. To address these concerns, innovative methodologies such as the chain of thoughts (COT) \cite{wei2022chain} reasoning and Retrieval-Augmented Generation (RAG) \cite{cai2022recent} have been developed. These approaches significantly enhance the fidelity of the models' responses. In particular, RAG represents a groundbreaking advancement by merging the powerful reasoning capabilities of foundation models (FM) with up-to-date external information. This is achieved by augmenting the input prompt with pertinent information retrieved from a comprehensive and up-to-date database. Such a process ensures that the model's generations are not only creative and contextually aware but also grounded in solid, verifiable evidence. Consequently, this approach markedly improves the accuracy and relevance of the results, mitigating some of the earlier concerns associated with large language models. The integration of external databases into the generative process ensures that the output of these models remains both innovative and anchored in reality.

\section{Method}

Our method allows the user to input any image of interest for geolocalization. Subsequently, the image is processed by the query and retrieval module, wherein the locations of the most similar and most dissimilar images are extracted. The image, along with these two sets of locations, is then fed into a multi-modality model. Finally, the geolocalization result is displayed as an interactive map, which can be explored via a web interface. We will explain each step in the following sections.
\begin{figure}[!b]
  \centering
  \includegraphics[width=0.5\columnwidth]{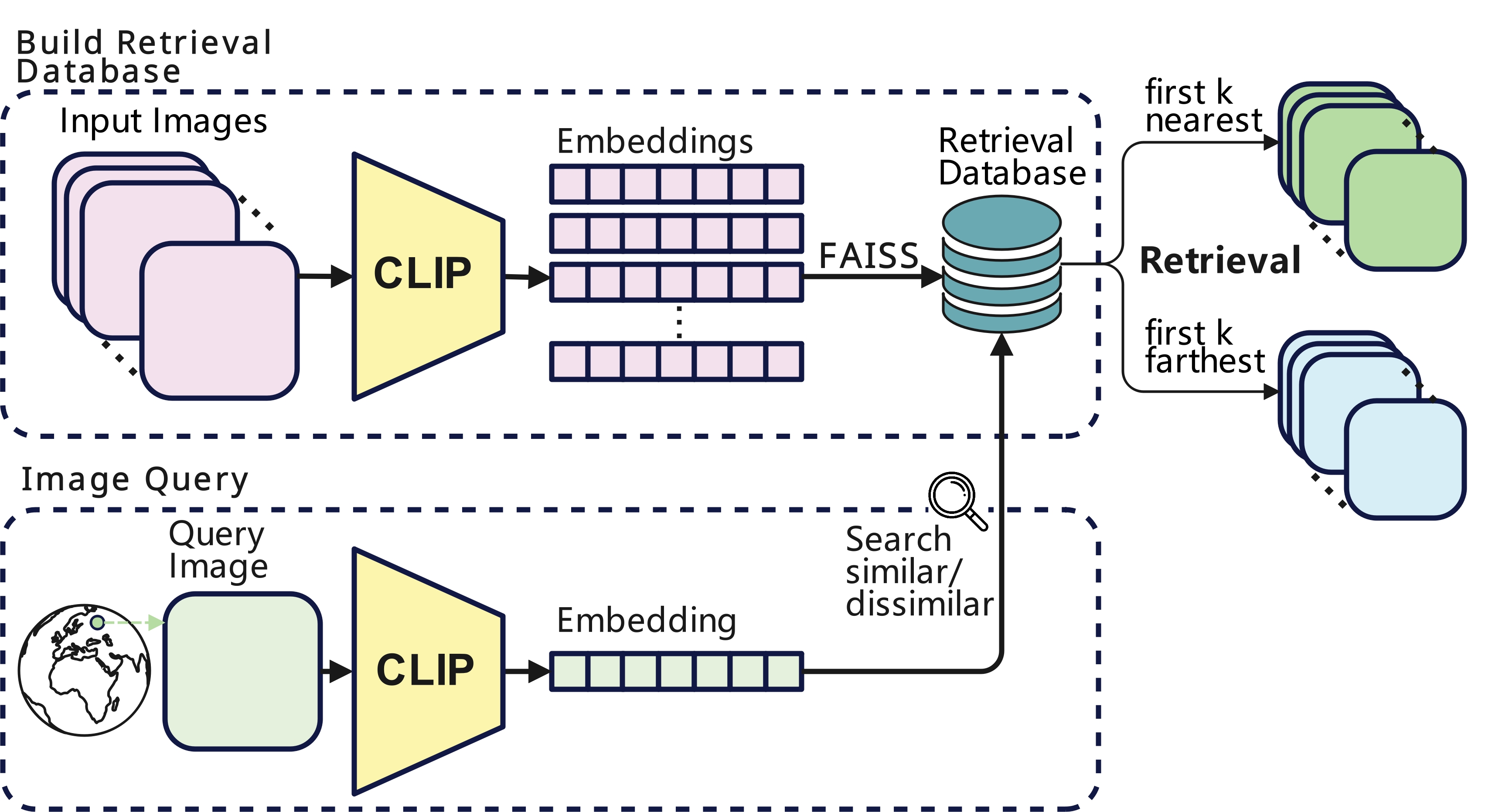}
  \caption{The retrieval and neighbor search pipeline.}
  \label{figure2}
\end{figure}

\subsection{Construction of the Image-Location  Database}

The core of the retrieval-based image localization system lies in how images are encoded into the database and how the nearest neighbor search is performed. Here, We utilize the CLIP model \cite{radford2021learning} for feature encoding and employ FAISS for the storage of resulting embeddings (Figure \ref{figure2}).

\textbf{CLIP-Based Feature Encoding.} The CLIP model, which is widely used as a fundamental representation model for various downstream tasks \cite{khan2022transformers}, is introduced to generate semantic embeddings of images. In particular, it accepts the query image and outputs the semantic embeddings of the outputs, which encapsulate rich information about the image in a condensed vector space. Utilizing the MediaEval Placing Tasks 2016 (MP-16) dataset \cite{larson2017benchmarking}, we have constructed a database encompassing over four million image embedding-location pairs, providing comprehensive coverage of the Earth's surface.

\textbf{Efficient Nearest Neighbor Search in the Vector Database.} Once image embeddings are generated, it becomes crucial to store them in an efficient manner to facilitate effective search operations. To address the challenge, we use FAISS, a vector-based data storage system \cite{johnson2019billion}, which utilizes flat indexes and GPU parallel computation techniques to enhance efficiency. Then, to find the nearest neighbor for the query image, we propose using the inner product of the image embedding provided by the CLIP as the measurement.
The underlying principle is straightforward: a higher inner product value signifies a greater level of similarity, and vice versa. This system allows us to generate an arbitrary number of nearest neighbors with ease. 

Moreover, we posit that identifying images most dissimilar to the query image (positive neighbors) can also contribute to ruling out implausible locations, as they usually represent scenes that are geographically distant from the query image. This ``negative neighbors search'' is executed by finding the farthest neighbors for the negative query embedding. Upon completion of this search, the locations of both positive and negative neighbors are integrated into the subsequent step of our process.

\subsection{Generate locations with augmented prompt} 

Current multi-modality foundation models, such as GPT-4V and LLaMA, accept input from both images and text to generate responses. In our approach, we conceptualize the task of image geolocalization as a text generation task. Specifically, we prompt these foundation models to provide the precise latitude and longitude corresponding to a given image. We enhance the input prompt with additional information derived from our retrieval of similar and dissimilar locations (Figure \ref{figure3}). The similar images' coordinates and dissimilar images' coordinates will be appended to the text prompt as anchor information.

\section{Experiments}
\subsection{Datasets and Evaluation Details}

We build our search database using the MediaEval Placing Tasks 2016 (MP-16) dataset \cite{larson2017benchmarking}, encompassing 4.72 million geotagged images from Flickr\footnote{https://www.flickr.com/}. The performance of our model is evaluated using Im2GPS3k \cite{vo2017revisiting} and YFCC4k \cite{vo2017revisiting} datasets. We compute the geodesic distance between the predicted and actual geographical coordinates for each test image and quantify the proportion of these predictions that align with set distance thresholds (1km, 25km, 200km, 750km, and 2500km). In terms of multi-modality models, our focus is on GPT-4V and LLaVA, selected for their availability and superior performance. It's noteworthy that our framework is designed for flexibility, allowing for seamless integration of the latest model releases as they become available.

\subsection{Results}

\begin{table}[!h]
\centering 

\resizebox{\columnwidth}{!}
{
\begin{tabular}{c | c || c c c c c}
\hline
\multirow{3}{*}{\bf Dataset} & \multirow{3}{*}{\bf \centering Method} & \multicolumn{5}{c}{\bf Distance ($a_r$ [\%] @ km)} \\ 

& & \multirow{1}{1.4 cm}{\tt \bf \centering Street} & \multirow{1}{1.4 cm}{\tt \bf \centering City} & \multirow{1}{1.4 cm}{\tt \bf \centering Region} & \multirow{1}{1.4 cm}{\tt \bf \centering Country} & \multirow{1}{1.4 cm}{\tt \bf \centering Continent} \\ 

& & \bf $1$ km & \bf $25$ km & \bf $200$ km & \bf $750$ km & \bf $2500$ km \\

\hline
\hline

\multirow{9}{1.2 cm}{\tt \centering \bf Im$2$GPS\\$3$k\\ \cite{vo2017revisiting}} & [L]kNN, $\sigma$ = $4$ \cite{vo2017revisiting} & 7.2 & 19.4 & 26.9 & 38.9 & 55.9 \\

& PlaNet \cite{weyand2016planet} & 8.5 & 24.8 & 34.3 & 48.4 & 64.6 \\

& CPlaNet \cite{seo2018cplanet} & 10.2 & 26.5 & 34.6 & 48.6 & 64.6 \\

& ISNs (M, f, S$_3$) \cite{muller2018geolocation} & 10.1 & 27.2 & 36.2 & 49.3 & 65.6 \\ 

& Translocator \cite{Pramanick22} &  11.8 & 31.1 & 46.7 & 58.9 & 80.1 \\

& GeoGuessNet \cite{clark2023we} & 12.8  & 33.5 & 45.9  & 61.0 & 76.1 \\

& GeoCLIP \cite{cepeda2023geoclip} & 14.11  & 34.47 & 50.65 & 69.67 & 83.82 \\

& \textbf{Img2Loc(LLaVA)} & 7.98  & 23.37 & 29.94  & 40.11 & 51.12 \\

&  \textbf{Img2Loc(GPT4V)} & \textbf{17.10}  & \textbf{45.14} & \textbf{57.87}  & \textbf{72.91} & \textbf{84.68} \\

\hline 
\hline

\multirow{8}{1.2 cm}{\tt \centering \bf YFCC\\$4k$\\ \cite{vo2017revisiting}} & [L]kNN, $\sigma$ = $4$ \cite{vo2017revisiting} & 2.3 & 5.7 & 11.0 & 23.5 & 42.0 \\

& PlaNet \cite{weyand2016planet} & 5.6 & 14.3 & 22.2 & 36.4 & 55.8 \\

& CPlaNet \cite{seo2018cplanet} & 7.9 & 14.8 & 21.9 & 36.4 & 55.5\\

& ISNs (M, f, S$_3$) \cite{muller2018geolocation} & 6.5 & 16.2 & 23.8 & 37.4 & 55.0 \\

& Translocator \cite{Pramanick22} & 8.4 & 18.6 & 27.0 & 41.1  & 60.4 \\

& GeoGuessNet\cite{clark2023we} & 10.3  & 24.4 & 33.9  & 50.0 & 68.7\\

& GeoCLIP\cite{cepeda2023geoclip} & 9.59  & 19.31 & 32.63  & 55.0 & 74.69\\

& \textbf{Img2Loc(LLaVA)} & 7.93  & 14.20 & 19.51  & 29.98 & 39.72 \\

& \textbf{Img2Loc(GPT4V)} & \textbf{14.11}  & \textbf{29.57} & \textbf{41.40}  & \textbf{59.27} & \textbf{76.88} \\

\hline 
\hline

\end{tabular}}
\caption{Geo-localization accuracy of the proposed method compared to previous methods, across two baseline datasets.} 

\label{tab:results_1}
\end{table}

The data presented in Table \ref{tab:results_1} demonstrates that our methods outperform previous classification and retrieval methods across all granularity levels on both tested datasets. On the Im2GPS3k dataset, we have achieved significant improvements over the prior top-performing method, GeoCLIP, without ever training any of the models on geo-tagged data (MP-16 dataset~\cite{larson2017benchmarking}) The improvements are +2.89\%, +10.67\%, +7.22\%, +3.24\%, and +0.86\% at the 1km, 25km, 200km, 750km, and 2500km thresholds, respectively. Furthermore, on the YFCC4k dataset, our method surpasses the previous best model, GeoGuessNet, by margins of +3.81\%, +5.17\%, +7.5\%, +9.27\%, and +8.18\% for the same respective distance thresholds.

\section{Conclusion}
In our study, we present Img2Loc, a cutting-edge system that harnesses the power of multi-modality foundation models and integrates advanced image-based information retrieval techniques for image geolocalization. Our approach has demonstrated evidently-improved performance when compared to existing methods. We envision Img2Loc as a compelling example of leveraging modern foundation models to address complex problems in a streamlined and effective manner.

\bibliographystyle{unsrt}  
\bibliography{references}

\end{document}